\documentclass[letterpaper]{article} 
\usepackage{aaai2026}  
\usepackage{times}  
\usepackage{helvet}  
\usepackage{courier}  
\usepackage[hyphens]{url}  
\usepackage{graphicx} 
\usepackage{natbib}  
\usepackage{caption} 
\usepackage{algorithm}
\usepackage{algorithmic}
\usepackage{amsmath}
\usepackage{amssymb}
\usepackage{bm}
\usepackage{booktabs}
\usepackage{multirow}
\usepackage[capitalize]{cleveref}
\crefname{section}{Sec.}{Secs.}
\Crefname{section}{Section}{Sections}
\Crefname{table}{Table}{Tables}
\crefname{table}{Tab.}{Tabs.}
\usepackage{newfloat}
\usepackage{listings}
\DeclareCaptionStyle{ruled}{labelfont=normalfont,labelsep=colon,strut=off} 
\floatstyle{ruled}
\newfloat{listing}{tb}{lst}{}
\floatname{listing}{Listing}
\nocopyright

\title{Light-IF: Endowing LLMs with Generalizable Reasoning via Preview and Self-Checking for Complex Instruction Following}
\author{
    Chenyang Wang\textsuperscript{\rm 1}\thanks{This work was done during the author's internship at 360.}\equalcontrib, Liang Wen\textsuperscript{\rm 2}\equalcontrib, Shousheng Jia\textsuperscript{\rm 2}, Xiangzheng Zhang\textsuperscript{\rm 2}\thanks{Corresponding Authors.}, Liang Xu\textsuperscript{\rm 3}\footnotemark[\value{footnote}]
}
\affiliations{
    \textsuperscript{\rm 1}Harbin Institute of Technology,
    \textsuperscript{\rm 2}Qiyuan Tech,
    \textsuperscript{\rm 3}CLUE\\

     zhangxiangzheng@360.cn, contact@superclue.ai
}

\usepackage{bibentry}

\begin{document}

\maketitle

\begin{abstract}
While advancements in the reasoning abilities of LLMs have significantly enhanced their performance in solving mathematical problems, coding tasks, and general puzzles, their effectiveness in accurately adhering to instructions remains inconsistent, particularly with more complex directives. Our investigation identifies lazy reasoning during the thinking stage as the primary factor contributing to poor instruction adherence. To mitigate this issue, we propose a comprehensive framework designed to enable rigorous reasoning processes involving preview and self-checking, essential for satisfying strict instruction constraints. 
Specifically, we first generate instructions with complex constraints and apply a filtering process to obtain valid prompts, resulting in three distinct prompt datasets categorized as hard, easy, and pass. Then, we employ rejection sampling on the pass prompts to curate a small yet high-quality dataset, enabling a cold-start initialization of the model and facilitating its adaptation to effective reasoning patterns.
Subsequently, we employ an entropy-preserving supervised fine-tuning (Entropy-SFT) strategy coupled with token-wise entropy-adaptive (TEA-RL) reinforcement learning guided by rule-based dense rewards. This approach encourages the model to transform its reasoning mechanism, ultimately fostering generalizable reasoning abilities that encompass preview and self-checking. Extensive experiments conducted on instruction-following benchmarks demonstrate remarkable performance improvements across various model scales. Notably, our Light-IF-32B model surpasses both larger open-source models such as DeepSeek-R1 and closed-source models like Doubao-1.6.
\end{abstract}

\begin{links}
    \link{Model}{https://huggingface.co/qihoo360/Light-IF-32B}
\end{links}


\section{Introduction}

\begin{figure}[t]
  \centering
  \includegraphics[width=1.0\linewidth]{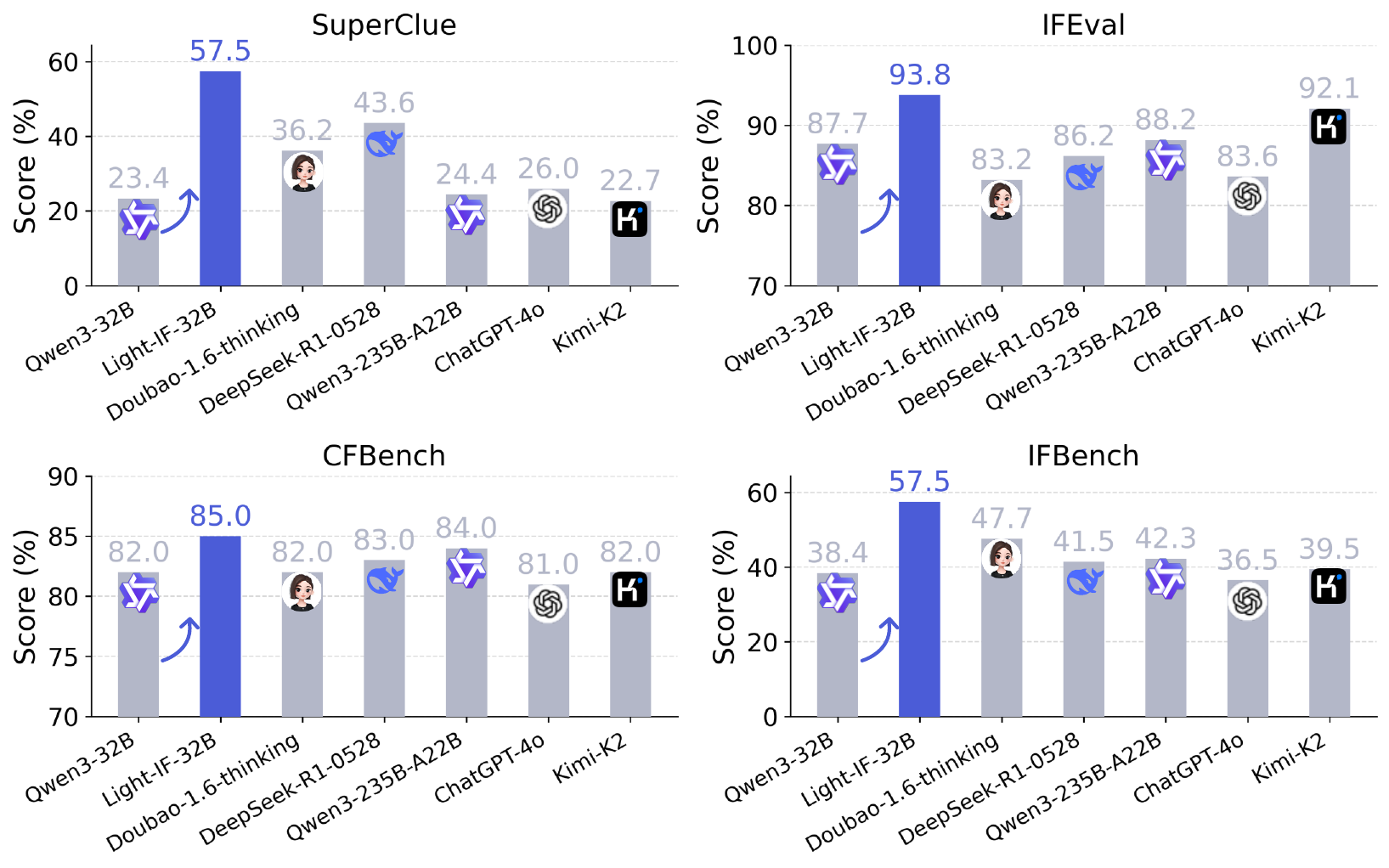}
  \caption{Main results of models on the instruction following benchmarks.}
  \label{fig:benchmark}
\end{figure}

\begin{figure*}[t]
  \centering
  \includegraphics[width=0.8\linewidth]{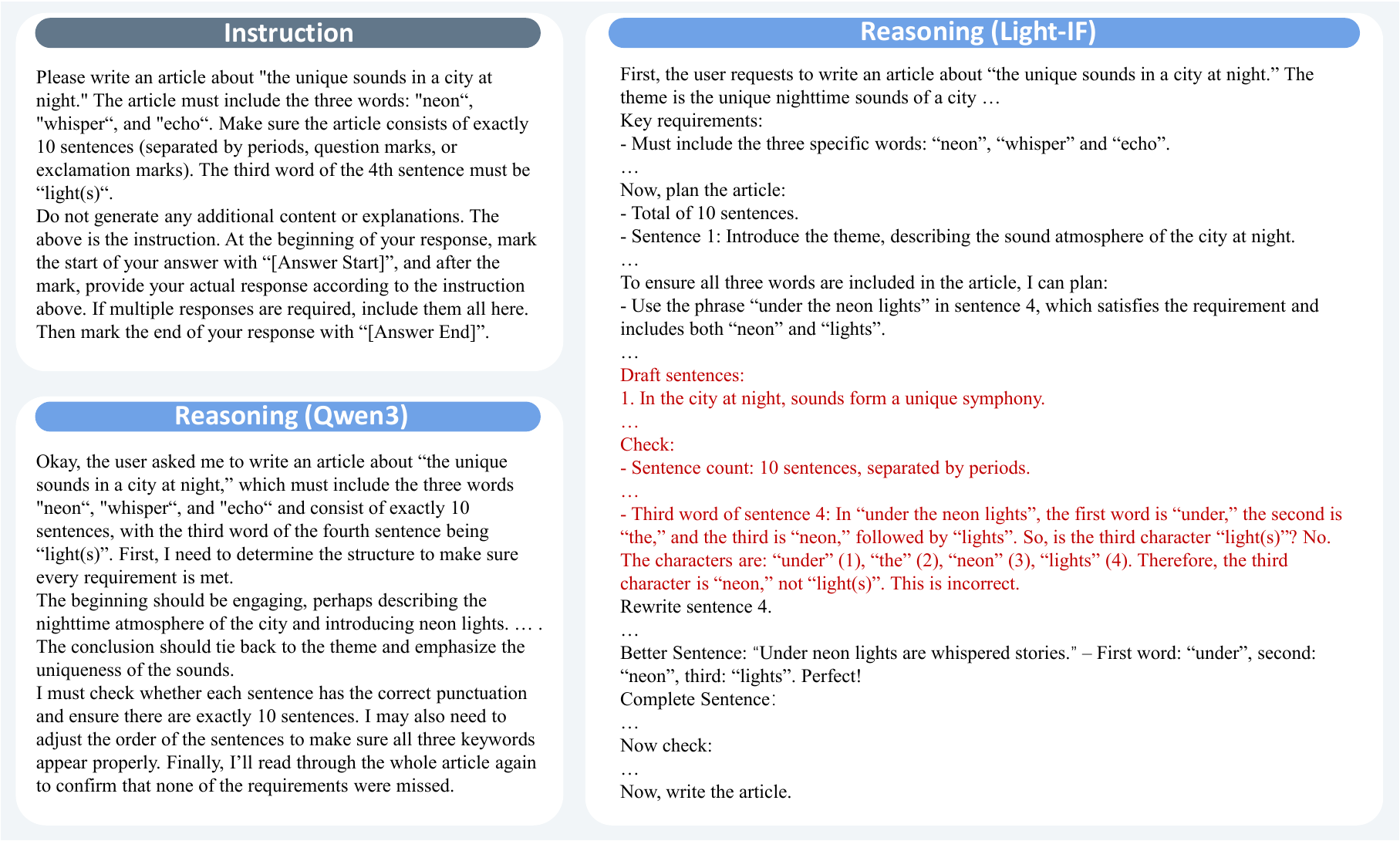}
  \caption{An example of different thinking patterns between Qwen3-32B and Light-IF-32B. Reasoning refers to the content between \textless think\textgreater{} and \textless /think\textgreater{}. The original text is in Chinese and translated here for readability.}
  \label{fig:example}
\end{figure*}

Instruction following~\citep{leike2018scalable,wei2021finetuned,lou2023comprehensive} is a fundamental capability of large language models (LLMs), marking their transition from mere next-token predictors to practical and reliable assistants. Models adept at instruction following can generate controlled outputs, aligning closely with human intentions and proving beneficial across diverse tasks~\citep{ouyang2022training,zhou2023instruction,zhang2024instructiontuninglargelanguage}. Conversely, models that fail to accurately interpret or adhere to instructions become unreliable, severely limiting their applicability in real-world domains such as healthcare~\citep{singhal2023large,cascella2023evaluating,singhal2025toward}, autonomous driving~\citep{ding2023hilm,shao2024lmdrive,cui2024large}, and agent-based systems~\citep{wang2024executable,team2025kimi,luo2025large}. 

Numerous benchmarks have been proposed to evaluate instruction following abilities~\citep{jing2023followeval,zhou2023instruction,xu2023superclue,jiang2023followbench,he2024can,wen2024benchmarking,li2024fb,zhang2024cfbench,qin2024infobench,pyatkin2025generalizing}. For instance, IFeval~\citep{zhou2023instruction} introduced verifiable atomic instructions to construct prompts. SuperCLUE~\citep{xu2023superclue} is a Chinese benchmark that scores exact compliance to strict, constraint-heavy instructions on complex tasks. CFBench~\citep{zhang2024cfbench} highlighted realistic multi-constraint compliance, featuring the broadest constraint coverage and a model-based scoring scheme. On top of IFeval, IFBench~\citep{pyatkin2025generalizing} introduced verifiable, out-of-domain constraints to reveal overfitting issues. \cref{fig:benchmark} illustrates the evaluation performance of several mainstream LLMs on these benchmarks, exposing need for further improvements.

Previous studies~\citep{sun2024conifer,wang2024direct,zhang2024iopo,dong2024self,xu2024wizardlm,zhao2024tree,huang2025musc,ren2025step} primarily enhanced complex instruction-following capabilities during post-training via SFT or direct preference optimization (DPO) on collected or synthetic instructions. For example, \citet{sun2024conifer} leveraged GPT-4 to categorize ShareGPT prompts into five difficulty levels, subsequently fine-tuning models through a progressively challenging curriculum. \citet{huang2025musc} created coarse and fine-grained contrastive pairs from the model itself, conducting multi-granularity self-contrastive DPO, while \citet{ren2025step} auto-generated soft-constraint data, and then trained models using DPO under a few-to-many constraints curriculum. Although effective, these methods heavily rely on extensive supervised data covering a broad spectrum of instructions, posing significant data collection challenges.

Recent advancements in reasoning LLMs~\citep{guo2025deepseek,yang2025qwen3,ye2025limo,wen2025light,wang2025beyond,bercovich2025llama,abdin2025phi,chen2025towards,deepscaler2025,sun2025tinyr1} demonstrate impressive generalization capabilities of their reasoning processes across various tasks, even when trained on limited task domains like mathematics and coding. Inspired by these developments, we resort to eliciting effective reasoning using a relatively small dataset to address complex instruction-following tasks. Our core intuition is that when an LLM adopts an effective reasoning strategy for a limited set of instructions, this strategy generalizes to unseen instructions with varying constraints and intentions. To begin with, we examine the behavior of recent reasoning LLMs, discovering a prevalent lazy reasoning pattern when confronted with complex instructions, as exemplified in \cref{fig:example}. This ineffective reasoning mode, characterized by simply restating instructions without genuine checking for compliance, hinders the model from strictly following the instructions of users. 

We propose a comprehensive framework to address this issue. First, we generate distinct instruction sets categorized as hard, easy, and pass, each with carefully controlled difficulty levels. Subsequently, we conduct Zero-RL training on the lazy-thinking model to incentivize effective reasoning behaviors. Leveraging the Zero-RL model and optionally external APIs, we perform rejection sampling to obtain thousands of high-quality responses with preview and self-checking, which serve as the cold-start dataset for the base model. Finally, we apply Entropy-SFT for cold-start initialization and subsequently train with TEA-RL, equipping LLMs with generalizable reasoning for IF tasks. The main contributions of this work are summarized as follows:
\begin{itemize}
\item We identify and characterize the lazy reasoning pattern prevalent in current reasoning LLMs when handling complex constraints.
\item We introduce an effective framework incorporating data collection, cold-start, and reinforcement learning stages, enabling generalizable and effective reasoning patterns characterized by previewing and self-checking mechanisms for complex instruction-following tasks.
\item We propose innovative techniques to effectively control entropy during both SFT and RL stages, specifically entropy-preserving SFT and RL with token-wise entropy-adaptive regularization.
\item Extensive experiments across multiple instruction-following benchmarks demonstrate superior performance of our Light-IF models.
\end{itemize}


\begin{figure}[t]
  \centering
  \includegraphics[width=1.0\linewidth]{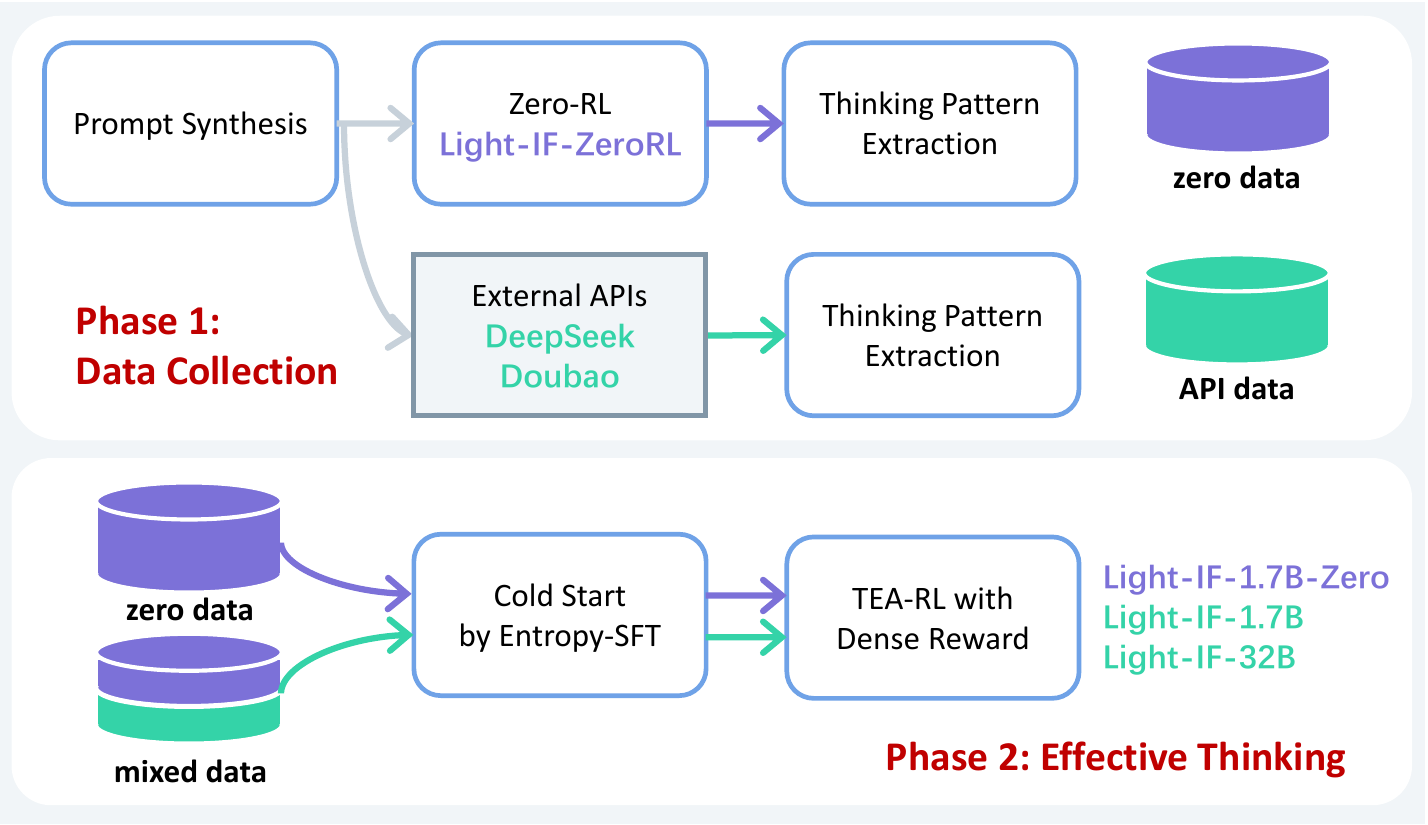}
  \caption{The overall framework of the proposed method.}
  \label{fig:framework}
\end{figure}

\section{Methods}
The proposed framework for enabling generalizable reasoning comprises five components: prompt synthesis, zero-RL from the lazy-thinking model, thinking pattern extraction, entropy-preserving SFT, and TEA-RL with dense rewards. The overall framework is depicted in \cref{fig:framework}.

\subsection{Hardness-aware Prompt Synthesis}

\begin{figure}[t]
  \centering
  \includegraphics[width=1.0\linewidth]{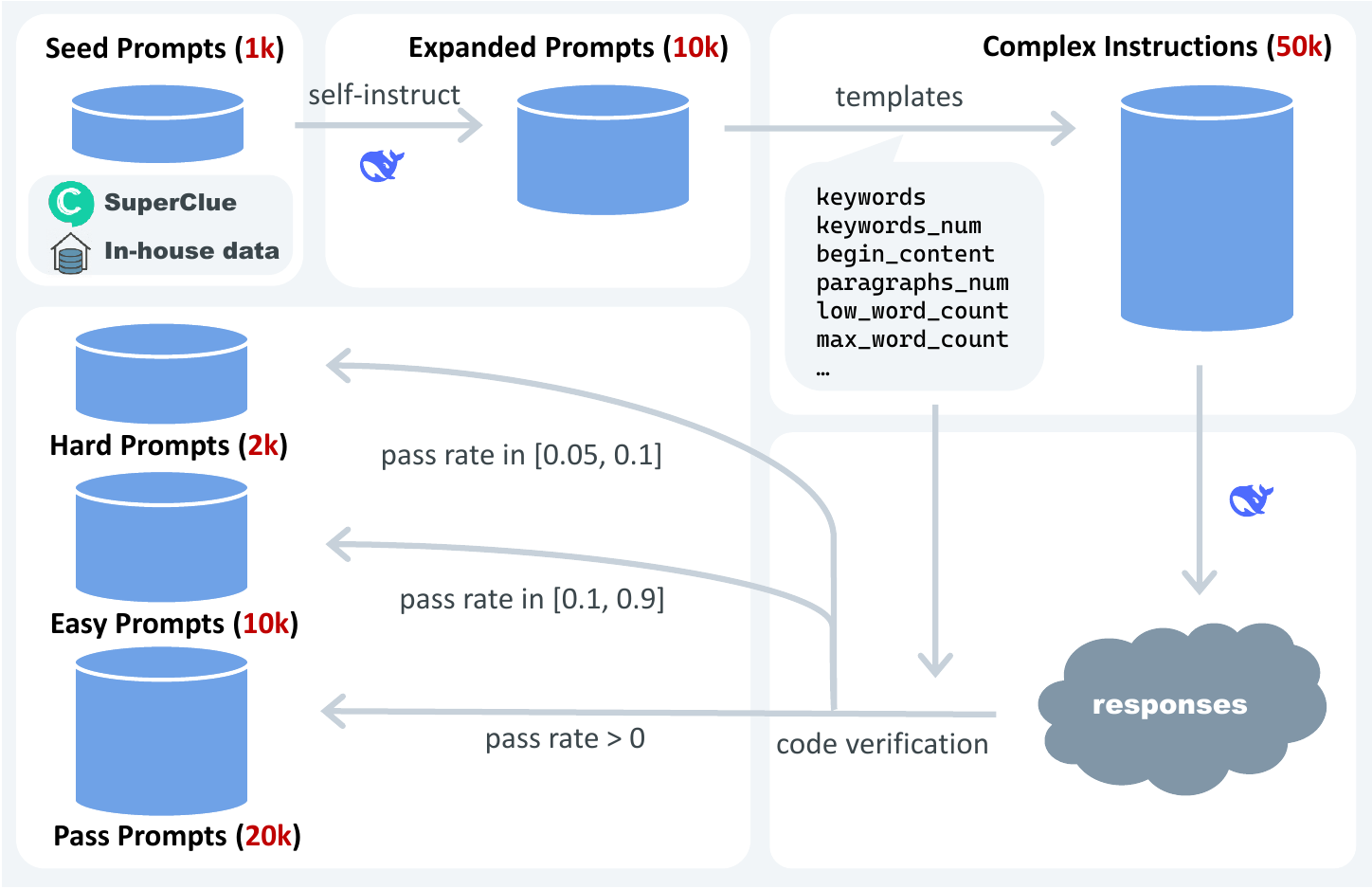}
  \caption{Pipeline of hardness-aware prompt synthesis.}
  \label{fig:data}
\end{figure}

Our pipeline for synthesizing hardness-aware prompts consists of four main steps: seed prompt collection, prompt expansion, construction of complex constraints, and prompt filtering. The overall pipeline is illustrated in \cref{fig:data}.

\textbf{Seed Prompt Collection}. Seed prompts originate from two primary sources: historical evaluation data from SuperClue~\citep{xu2023superclue} (100+ Chinese prompts collected before October 2024) and our in-house evaluation datasets (200+ English prompts and 600+ Chinese prompts). \textbf{Prompt Expansion}. We employ the Self-Instruct~\citep{wang2022self} methodology to expand seed prompts with simple instructions, resulting in an expanded set of 10,000 prompts.
\textbf{Construction of Complex Constraints}. In this stage, we exclusively incorporate limited verifiable constraints. Specifically, following AutoIF~\citep{dong2024self}, we define an instruction template structured as a dictionary, including keys such as keyword frequency, word count, paragraph count, among others. The values for each instruction template are randomly sampled. By applying five different templates to each simple instruction, we generate 50,000 prompts with complex instructions.
\textbf{Prompt Filtering}. To eliminate invalid prompts, we utilize an efficient LLM to generate ten outputs per prompt. Due to the verifiable nature of these instructions, we discard prompts whose outputs consistently fail code verification. This initial filtering process results in approximately 20,000 valid prompts, designated as pass prompts. Subsequently, we create two distinct datasets categorized by their difficulty levels: an easy prompt dataset with pass ratios ranging from $[0.1, 0.9]$ and a hard prompt dataset with pass ratios ranging from $[0.05, 0.1]$.

\subsection{Zero-RL from Lazy-Thinking Model}
\begin{figure}[t]
  \centering
  \includegraphics[width=1.0\linewidth]{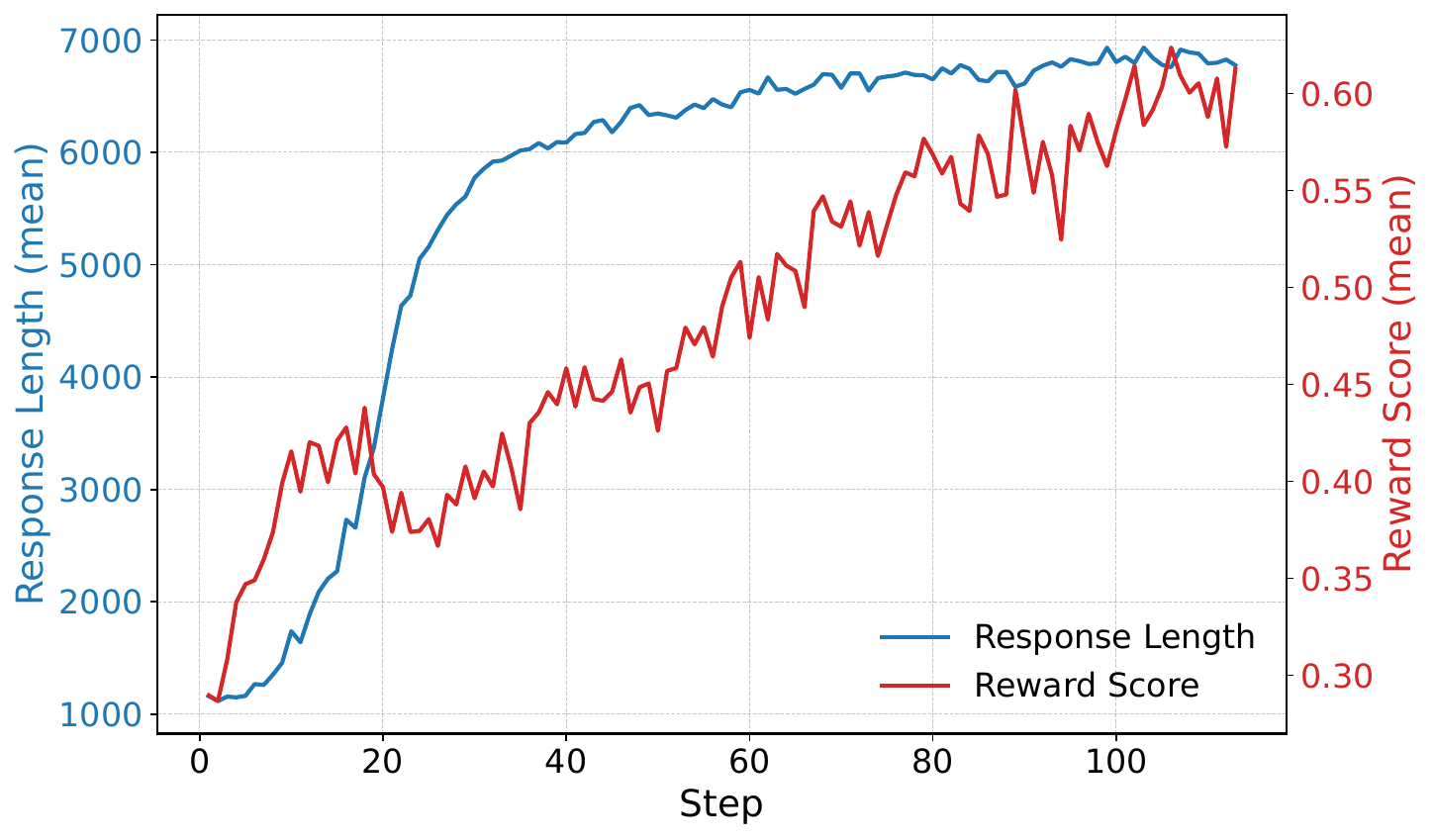}
  \caption{Response length and $R_c$ during Zero-RL training.}
  \label{fig:Zero}
\end{figure}

To incentivize the effective reasoning pattern, we adopt R1-Zero-style reinforcement learning (Zero-RL) to post-train the lazy-thinking model. The key challenge in Zero-RL lies in the design of rewards. Specifically, the rewards in Zero-RL consist of two components: (1) correctness score $R_{c}$ and (2) length reward $R_l$. The correctness reward (and other training designs) will be detailed in Section TAE-RL. In this section, we focus specifically on the length reward, defined as follows:
\begin{align}
R_l = 
\begin{cases}
-2, & \text{if } L \geq L_{max}, \\
 2 \cdot R_c \cdot \gamma(L), & \text{if } R_c \geq 0.2 \text{ and } L < L_{max}, \\
-\gamma(L), & \text{if } R_c < 0.2 \text{ and }L < L_{max}.
\end{cases}
\end{align}
where $\gamma(L) = 0.5 \left( 1 - \cos\left( \pi \cdot \frac{L}{L_{max}} \right) \right)$. The length reward encourages correct and sufficiently long responses to explicitly mitigate lazy thinking. Simultaneously, it penalizes excessively long responses as well as incorrect and overly verbose outputs. The relationship between response length and correctness score $R_c$ during Zero-RL is illustrated in \cref{fig:Zero}. Apart from initial fluctuations, response length exhibits a strong positive correlation with correctness scores. The initial reduction also highlights the difficulty of using the correctness reward alone, where the model doesn't prefer longer responses when the length achieves 3,000, making it hard to learn effective thinking pattern. Moreover, we observed the emergence of preview and self-checking behaviors, aligning well with our expectations.

\subsection{Thinking Pattern Extraction}
To gather samples exhibiting effective reasoning patterns for cold-start, we design a data filtering pipeline for model responses to pass prompts (20K). The pipeline includes correctness check, thinking check, and fluency check (detailed in the Appendix C). After filtering, we collect the top 2,000 high-quality cold-start samples, evenly split between Chinese and English.

Depending on the use of external APIs, we construct two distinct cold-start datasets:
1) \textbf{zero data}: the acquirement of zero data doesn't rely on external guidance, where all responses are generated by the Zero-RL model.
2) \textbf{mixed data}: we mix responses from Zero-RL model (covering all pass prompts) and external APIs (DeepSeek-R1 for Chinese subset and Doubao-1.6 for English subset). These APIs demonstrate stable and effective reasoning patterns in certain scenarios, making them suitable candidates for sampling. The mixed responses are then filtered using the same pipeline described above.
We train two separate models using these distinct cold-start datasets for the following reasons:
1) the Zero-RL-based cold start showcases the viability of eliciting effective reasoning patterns entirely from a lazy-thinking model, without relying on external guidance—offering a practical and self-contained solution for the community.
2) by leveraging the strong external APIs for cold-starting, we obtain a model with SOTA performance, offering the community a powerful open-source model that requires significantly fewer parameters.


\subsection{Entropy-Preserving SFT for Cold-Start}

\begin{figure}[t]
  \centering
  \includegraphics[width=1.0\linewidth]{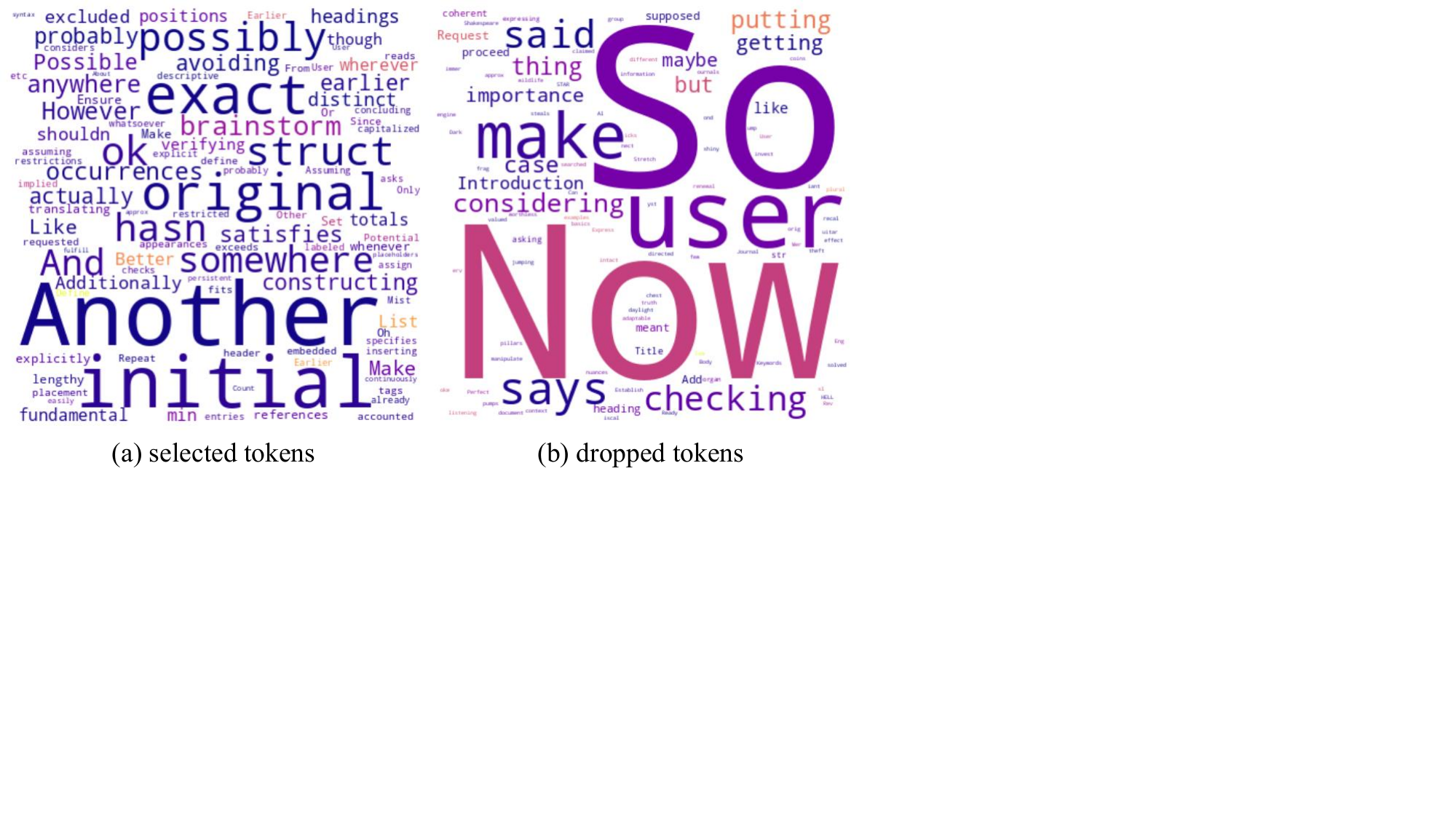}
  \caption{Left: Top-100 tokens with high selection rank at the beginning of the cold-start stage. Right: Top-100 tokens with low selection rank at the end of the cold-start stage. Larger size indicates larger frequency and brighter color indicates higher rank.}
  \label{fig:token1}
\end{figure}

Entropy~\citep{cui2025entropy,wang2025beyond,yu2025dapo,fu2025srft} serves as a metric for exploration in LLMs during reinforcement learning and is considered a crucial indicator of their final performance. SFT aims to minimize the mean cross-entropy loss for each training sample, inevitably causing entropy reduction on the training data, given the large number of model parameters and limited training samples. Furthermore, we observe that this entropy reduction phenomenon generalizes to untrained samples as well, decreasing the initial entropy compared to the base model and negatively impacting subsequent reinforcement learning stages. Therefore, we propose entropy-preserving SFT to mitigate entropy loss during the cold-start phase.

Specifically, inspired by \citet{wang2025beyond}, we think that not all tokens are necessary for loss computation during training. Thus, during the cold-start stage, we selectively choose only $r\%$ of tokens for each training step based on prediction entropy and cross-entropy loss. Formally, the prediction distribution for token $o_t$ of sample $s_i$ is defined as:
\begin{equation}
\bm{p^i_t}=\pi_\theta(\cdot|q^i,o^i_{<t}),
\end{equation}
where $\pi_\theta$ represents the LLM parameterized by $\theta$, $q^i$ is the input query of sample $s_i$, and $o^i_{<t}$ denotes outputs prior to token $o_t$. The prediction entropy is defined as $H_t^i=-\sum_j(\bm{p^i_t})_j \log (\bm{p^i_t})_j$, and the cross-entropy loss is defined as $\mathrm{NLL}_t^i=-\log \bm{p^i_t}(o_t)$. For a batch of samples $B$, the selected tokens $T_{sel}$ are determined as follows:
\begin{align}
    &T_{all} = \{o_t|o_t\in s_i \land s_i \in B\}, \\ \nonumber
    &S = \{\mathrm{NLL}_t-\alpha H_t|o_t\in T_{all}\}, \\ \nonumber
    &T_{sel} = \{o_t|o_t\in T_{all}\land \mathrm{NLL}_t-\alpha H_t>\mathrm{Quantile}_r(S)\}.
\end{align}
The Entropy-SFT loss is then computed on selected tokens:
\begin{equation}
    L_{Entropy-SFT} =\frac{1}{|T_{sel}|}\sum_{o_t\in T_{sel}} -\log\bm{p}_t(o_t).
\end{equation}

This token selection strategy serves two purposes:
1) the first component prioritizes tokens with high negative log-likelihood, representing new information introduced by cold-start data not captured by the baseline model. As illustrated at the left of \cref{fig:token1}, tokens selected at the beginning typically relate to preview and checking pattern. Examples include sentences like "\textit{Another} approach for the third bullet: [preview for the third bullet]" and "Wait, but in the \textit{initial} response, [check constraints for responses]."
2) the second component accounts for token entropy, assigning lower selection ranks to high-entropy tokens to preserve the model's inherent entropy at the end of the cold-start. As depicted at the right of \cref{fig:token1}, tokens excluded at the end of training are either confidence tokens with low NLL (\textit{e.g.} \textit{user} and \textit{plural}) or high entropy tokens (\textit{e.g.} \textit{Now} and \textit{considering}). These tokens are appropriately dropped to avoid ineffective training and to preserve entropy.

\subsection{TEA-RL with Dense Reward}
\begin{table}[t]
  \centering
  \small               
  \setlength{\tabcolsep}{4pt}   
  \begin{tabular}{lllc}
    \toprule
    \textbf{Group} & \textbf{Item} & \textbf{Constraint Type} & \textbf{Reward} \\ 
    \midrule
    \multirow{5}{*}{Keywords} 
      & Keyword 1 & $[n_{\min}, n_{\max}]\land n_{max}<5$ & 0.10 \\
      & Keyword 2 & $[n_{\min}, n_{\max}]\land n_{max}\geq 5$ & 0.20 \\
      & Keyword 3 & $\le n_{\min}$         & 0.05 \\
      & Keyword 4 & $\ge n_{\max}$         & 0.05 \\
      & Keyword 5 & $=n$                   & 0.10 \\ \midrule
    Paragraphs & — & $=n$  & 0.10 \\ 
    Sentences  & — & $=n$  & 0.20 \\ \midrule
    \multirow{4}{*}{Words} 
      & Word 1 & $[n_{\min}, n_{\max}]\land \Delta>50$ & 0.10 \\
      & Word 2 & $[n_{\min}, n_{\max}]\land \Delta\leq50$ & 0.20 \\
      & Word 3 & $\le n_{\min}$         & 0.05 \\
      & Word 4 & $\ge n_{\max}$         & 0.05 \\ \midrule
    Beginning & — & match    & 0.02 \\
    End       & — & match    & 0.02 \\ \midrule
    All   & — & satisfy  & +1.00 \\
    \bottomrule
  \end{tabular}
  \caption{Constraint types and corresponding rewards. If all constraints are met, the model receives an extra reward of 1.}
  \label{tab:constraints}
\end{table}

The RL stage encounters two primary challenges: sparse reward and entropy collapse. The sparse reward issue arises because, although final instruction-following outcomes are verifiable, the traditional binary reward signal provides extremely sparse feedback when the model initially struggles to fully satisfy complex constraints. Given the nature that complex instructions consist of multiple constraints, we adopt a strategy of rewarding partial constraint satisfaction individually, as illustrated in \cref{tab:constraints}. With this dense reward strategy, the reward curve exhibits a smoother and more rapid increase (detailed in Appendix E).

Entropy collapse refers to the phenomenon where the model’s entropy rapidly diminishes during training, significantly reducing exploration capability and resulting in suboptimal performance. A straightforward approach to address this problem is entropy regularization during the reinforcement learning phase. However, directly applying entropy regularization either leads to entropy explosion or fails to effectively halt entropy collapse. Building upon the theoretical insight that \textit{changes in policy entropy are driven by the covariance between action probability and logit changes, proportional to advantages when employing policy gradient algorithms~\citep{cui2025entropy}}, we propose the token-level entropy-adaptive regularization term. Formally, for a rollout batch, the covariance for tokens $T_{r}$ is calculated as:
\begin{align}
\mathrm{Cov}_t = &\left( \log \bm{p}_t(o_t) - \frac{1}{|T_{r}|} \sum_{o_j\in T_{r}} \log\bm{p}_t(o_j) \right) \nonumber \\ 
&\cdot \left( A(o_t) - \frac{1}{|T_{r}|} \sum_{o_j\in T_{r}} A(o_j) \right).
\end{align}
The token-level regularization term is then defined as:
\begin{equation}
    L_{TEA} = |T_{r}|\sum_{o_t\in T_{r}} \min\left(\frac{e^{\mathrm{Cov}_t/\tau}}{\sum_{o_j \in T_{r}}e^{\mathrm{Cov}_j/\tau}},\frac{c}{|T_{r}|}\right) H_t,
\end{equation}
where $\tau$ is the temperature and $c$ is the max coefficient value. Consequently, the total loss for TEA-RL is given by:
\begin{equation}
    L_{TEA-RL} = L_{GRPO} - \lambda L_{TEA},
\end{equation}
where $\lambda$ represents the regularization coefficient.

TEA-RL training is conducted in two stages, following an easy-to-hard curriculum, with the model first trained on easy prompts and subsequently on hard prompts. Notably, to demonstrate the generalization capability of our framework, we only perform RL on the Chinese subset of each dataset.

\section{Experiments}
In this section, we first present evaluation results on instruction-following benchmarks. Next, we examine the performance progression at each stage of the Zero-RL pipeline. Subsequently, we validate the effectiveness of specific design choices within our framework through ablation studies. Lastly, we analyze the entropy dynamics.

\begin{table*}[t]
\centering
\footnotesize
\setlength{\tabcolsep}{3pt}       
\begin{tabular}{lccccccccccccc}
\hline
\multirow{2}{*}{Model}        & \multirow{2}{*}{SuperClue} & \multicolumn{5}{c}{IFEval}                                                                                       & \multicolumn{4}{c}{CFBench}                                                               & \multicolumn{3}{c}{IFBench}                                        \\
                              &                            & LP                   & LI                   & SP                   & SI                   & AVG                  & CSR                  & ISR                  & PSR                  & AVG                  & PL                   & IL                   & AVG                  \\ \hline
\textit{Non-reasoning Models} & \multicolumn{1}{l}{}       & \multicolumn{1}{l}{} & \multicolumn{1}{l}{} & \multicolumn{1}{l}{} & \multicolumn{1}{l}{} & \multicolumn{1}{l}{} & \multicolumn{1}{l}{} & \multicolumn{1}{l}{} & \multicolumn{1}{l}{} & \multicolumn{1}{l}{} & \multicolumn{1}{l}{} & \multicolumn{1}{l}{} & \multicolumn{1}{l}{} \\
ChatGPT-4o                    & 0.260                      & 0.837                & 0.880                & 0.786                & 0.841                & 0.836                & 0.90                 & 0.72                 & 0.80                 & 0.81                 & 0.354                & 0.376                & 0.365                \\
Deepseek-v3-0324              & 0.306                      & 0.856                & 0.899                & 0.815                & 0.867                & 0.859                & 0.91                 & 0.76                 & 0.83                 & 0.83                 & 0.388                & 0.421                & 0.405                \\
Doubao-1.5-pro                & 0.285                      & 0.885                & 0.921                & 0.852                & 0.899                & 0.889                & 0.89                 & 0.71                 & 0.79                 & 0.80                 & 0.361                & 0.388                & 0.375                \\
Kimi-K2                       & 0.227                      & 0.917                & 0.944                & 0.895                & 0.929                & 0.921                & 0.91                 & 0.74                 & 0.82                 & 0.82                 & 0.378                & 0.412                & 0.395                \\ \hline
\textit{Reasoning Models}     &                            &                      &                      &                      &                      &                      &                      &                      &                      &                      &                      &                      &                      \\
Qwen3-1.7B                    & 0.081                      & 0.726                & 0.796                & 0.697                & 0.767                & 0.747                & 0.81                 & 0.56                 & 0.67                 & 0.68                 & 0.275                & 0.301                & 0.288                \\
Qwen3-32B                     & 0.234                      & 0.871                & 0.914                & 0.834                & 0.887                & 0.877                & 0.91                 & 0.74                 & 0.82                 & 0.82                 & 0.364                & 0.403                & 0.384                \\
Qwen3-235B-A22B               & 0.244                      & 0.884                & 0.916                & 0.839                & 0.887                & 0.882                & 0.92                 & 0.75                 & 0.84                 & 0.84                 & 0.408                & 0.438                & 0.423                \\
DeepSeek-R1-0528              & 0.436                      & 0.857                & 0.905                & 0.814                & 0.873                & 0.862                & 0.91                 & 0.74                 & 0.83                 & 0.83                 & 0.405                & 0.424                & 0.415                \\
Doubao-1.6-thinking           & 0.362                      & 0.825                & 0.881                & 0.775                & 0.847                & 0.832                & 0.90                 & 0.74                 & 0.82                 & 0.82                 & 0.469                & 0.485                & 0.477                \\ \hline
\textit{Our Models}           &                            &                      &                      &                      &                      &                      &                      &                      &                      &                      &                      &                      &                      \\
Light-IF-1.7B-Zero            & 0.232                      & 0.872                & 0.917                & 0.836                & 0.901                & 0.882                & 0.84                 & 0.61                 & 0.72                 & 0.72                 & 0.375                & 0.407                & 0.391                \\
Light-IF-1.7B                 & 0.299                      & 0.885                & 0.924                & 0.856                & 0.904                & 0.892                & 0.85                 & 0.61                 & 0.71                 & 0.72                 & 0.371                & 0.415                & 0.393                \\
Light-IF-32B                  & \textbf{0.575}             & \textbf{0.933}       & \textbf{0.956}       & \textbf{0.917}       & \textbf{0.945}       & \textbf{0.938}       & \textbf{0.93}        & \textbf{0.77}        & \textbf{0.85}        & \textbf{0.85}        & \textbf{0.565}       & \textbf{0.585}       & \textbf{0.575}       \\ \hline
\end{tabular}
  \caption{Evaluation results. LP: loose prompt, LI: loss instruction, SP: strict prompt, SI: strict instruction, PL: prompt level, IL: instruction level. Best are marked \textbf{bold} among all models.}
  \label{tab:eval}
\end{table*}

\subsection{Experimental Settings}

\begin{table*}[t]
\centering
\footnotesize
\setlength{\tabcolsep}{3pt}       
\begin{tabular}{lccccccccccccc}
\hline
\multirow{2}{*}{Model}            & \multirow{2}{*}{SuperClue} & \multicolumn{5}{c}{IFEval}                                                         & \multicolumn{4}{c}{CFBench}                                   & \multicolumn{3}{c}{IFBench}                      \\
                                  &                            & LP             & LI             & SP             & SI             & AVG            & CSR           & ISR           & PSR           & AVG           & PL             & IL             & AVG            \\ \hline
Qwen3-1.7B                        & 0.081                      & 0.726          & 0.796          & 0.697          & 0.767          & 0.747          & 0.81          & 0.56          & 0.67          & 0.68          & 0.275          & 0.301          & 0.288          \\
Qwen3-32B                         & 0.234                      & 0.871          & 0.914          & 0.834          & 0.887          & 0.877          & 0.91          & 0.74          & 0.82          & 0.82          & 0.364          & 0.403          & 0.384          \\
Light-IF-1.7B-ZeroRL              & 0.157                      & 0.843          & 0.886          & 0.804          & 0.855          & 0.847          & 0.82          & 0.59          & 0.68          & 0.70          & 0.327          & 0.355          & 0.341          \\
Light-IF-1.7B-EntSFT(ZR)          & 0.148                      & 0.841          & 0.888          & 0.797          & 0.860          & 0.847          & 0.84          & 0.59          & 0.70          & 0.71          & 0.350          & 0.388          & 0.369          \\
Light-IF-1.7B-EntSFT(ZR)-TEARL    & 0.175                      & 0.869          & 0.912          & 0.826          & 0.881          & 0.872          & \textbf{0.85} & \textbf{0.62} & 0.71          & \textbf{0.73} & 0.364          & 0.394          & 0.379          \\
Light-IF-1.7B-EntSFT(ZR)-TEARL-s2 & \textbf{0.232}             & \textbf{0.872} & \textbf{0.917} & \textbf{0.836} & \textbf{0.901} & \textbf{0.882} & 0.84          & 0.61          & \textbf{0.72} & 0.72          & \textbf{0.375} & \textbf{0.407} & \textbf{0.391} \\ \hline
\end{tabular}
  \caption{Zero-RL model series performance. LP: loose prompt, LI: loss instruction, SP: strict prompt, SI: strict instruction, PL: prompt level, IL: instruction level. Best are marked \textbf{bold} among 1.7B models.}
  \label{tab:zero}
\end{table*}

\begin{table*}[t]
\centering
\small
\setlength{\tabcolsep}{3pt}       
\begin{tabular}{cccc|ccccccccccccc}
\hline
\multirow{2}{*}{cold-start} & \multirow{2}{*}{RL stage} & \multirow{2}{*}{EntSFT} & \multirow{2}{*}{TEARL} & \multirow{2}{*}{Superclue} & \multicolumn{5}{c}{IFEval}                                                                                                                & \multicolumn{4}{c}{CFBench}                                                                                        & \multicolumn{3}{c}{IFBench}                                                       \\
                           &                           &                         &                        &                            & LP                        & LI                        & SP                        & SI                        & AVG                       & CSR                      & ISR                      & PSR                               & AVG                      & PL                        & IL                        & AVG                       \\ \hline
\multicolumn{1}{l}{}       & \multicolumn{1}{l}{}      & \multicolumn{1}{l}{}    & \multicolumn{1}{l|}{}  & 0.081                      & \multicolumn{1}{l}{0.726} & \multicolumn{1}{l}{0.796} & \multicolumn{1}{l}{0.697} & \multicolumn{1}{l}{0.767} & \multicolumn{1}{l}{0.747} & \multicolumn{1}{l}{0.81} & \multicolumn{1}{l}{0.56} & \multicolumn{1}{l}{0.67}          & \multicolumn{1}{l}{0.68} & \multicolumn{1}{l}{0.275} & \multicolumn{1}{l}{0.301} & \multicolumn{1}{l}{0.288} \\
$\checkmark$               &                           & $\checkmark$            &                        & 0.065                      & \multicolumn{1}{l}{0.828} & \multicolumn{1}{l}{0.881} & \multicolumn{1}{l}{0.784} & \multicolumn{1}{l}{0.850} & \multicolumn{1}{l}{0.836} & \multicolumn{1}{l}{0.84} & \multicolumn{1}{l}{0.59} & \multicolumn{1}{l}{\textbf{0.70}} & \multicolumn{1}{l}{0.71} & \multicolumn{1}{l}{0.328} & \multicolumn{1}{l}{0.350} & \multicolumn{1}{l}{0.339} \\
                           & $\checkmark$              &                         & $\checkmark$           & 0.157                      & 0.843                     & 0.886                     & 0.804                     & 0.855                     & 0.847                     & 0.82                     & 0.59                     & 0.68                              & 0.70                     & 0.327                     & 0.355                     & 0.341                     \\
$\checkmark$               & $\checkmark$              &                         &                        & 0.135                      & 0.810                     & 0.865                     & 0.762                     & 0.837                     & 0.819                     & 0.82                     & 0.55                     & 0.65                              & 0.67                     & 0.310                     & 0.334                     & 0.322                     \\
$\checkmark$               & $\checkmark$              & $\checkmark$            &                        & 0.153                      & 0.815                     & 0.872                     & 0.776                     & 0.842                     & 0.826                     & \textbf{0.85}            & 0.59                     & \textbf{0.70}                     & 0.71                     & 0.323                     & 0.361                     & 0.342                     \\
$\checkmark$               & $\checkmark$              &                         & $\checkmark$           & 0.166                      & 0.843                     & 0.893                     & 0.795                     & 0.859                     & 0.845                     & 0.83                     & 0.55                     & 0.66                              & 0.68                     & 0.310                     & 0.355                     & 0.333                     \\
$\checkmark$               & $\checkmark$              & $\checkmark$            & $\checkmark$           & \textbf{0.231}             & \textbf{0.880}            & \textbf{0.917}            & \textbf{0.845}            & \textbf{0.891}            & \textbf{0.883}            & \textbf{0.85}            & \textbf{0.60}            & \textbf{0.70}                     & \textbf{0.72}            & \textbf{0.344}            & \textbf{0.391}            & \textbf{0.368}            \\ \hline
\end{tabular}
  \caption{Ablation Study. LP: loose prompt, LI: loss instruction, SP: strict prompt, SI: strict instruction, PL: prompt level, IL: instruction level. Best are marked \textbf{bold}.}
  \label{tab:ablation}
\end{table*}

\paragraph{Benchmarks.} To evaluate the performance of various LLMs, we employ four distinct benchmarks: IFEval, CFBench, IFBench, and SuperCLUE-CPIF~\shortcite{CLUE2025SuperCLUE} (abbreviated to SuperCLUE in the remaining). We adopt DeepSeek-V3 as the judge model for CFBench and use greedy decoding on all benchmarks.

\paragraph{Baselines.} The baseline models include the Qwen3 series~\citep{yang2025qwen3} (Qwen3-1.7B, Qwen3-32B, and Qwen3-235B-A22B), DeepSeek series (DeepSeek-V3~\citep{liu2024deepseek} and DeepSeek-R1~\citep{guo2025deepseek}), Doubao series (Doubao-1.5-pro~\shortcite{Volcengine2025Doubao15}, Doubao-1.6-thinking~\shortcite{Volcengine2025Doubao16}), ChatGPT-4o~\shortcite{OpenAI2024GPT4o} and Kimi-K2~\citep{team2025kimi}. All comparison models are powerful, up-to-date, and actively deployed in production environments.

\paragraph{Implementation Details.} For cold-start training, we utilize the LLaMA-Factory~\citep{zheng2024llamafactory} with our modified Entropy-SFT. For the subsequent RL stage, we adopt VeRL~\citep{sheng2024hybridflow} integrated with our proposed TEA regularization. The hyperparameters $(r,\alpha)$ for Entropy-SFT and $(\tau, \lambda, c)$ for TEA-RL are set to values $(80, 0.8)$ and $(1.0, 0.05, 100)$, respectively. Further implementation details are provided in the Appendix B.

\subsection{Evaluation Results}
In this section, we evaluate the performance of our models against recent strong reasoning and non-reasoning models. Specifically, we select three model variants for evaluation: Light-IF-32B-EntSFT-TEARL-s2, Light-IF-1.7B-EntSFT-TEARL-s2, and Light-IF-1.7B-EntSFT(ZR)-TEARL-s2, where EntSFT denotes entropy-preserving SFT, ZR denotes cold-starting from zero data, and s2 indicates that the model has undergone two stages of RL training. For brevity, we refer to these models as Light-IF-32B, Light-IF-1.7B, and Light-IF-1.7B-Zero, respectively.

As demonstrated by the results in \cref{tab:eval}, our models equipped with effective reasoning patterns show remarkable improvements over their corresponding base models (Qwen3-1.7B and Qwen3-32B). Notably, Light-IF-32B achieves the highest performance among all evaluated models, surpassing the next-best models by 13.9, 1.7, 1.0, and 9.8 points on SuperClue, IFEval, CFBench, and IFBench, respectively. Moreover, Light-IF-1.7B demonstrates impressive performance despite having significantly fewer parameters, outperforming Qwen3-235B-A22B and Qwen3-32B on SuperClue and IFEval, and closely matching them on IFBench. Furthermore, Light-IF-1.7B-Zero exhibits competitive performance relative to Light-IF-1.7B, which highlights the potential of the base model to independently enhance reasoning capabilities without reliance on external APIs. Importantly, these evaluation outcomes underscore the generalizable effectiveness of the reasoning patterns learned by our models. Despite being trained on a limited set of synthetic constraints, our models generalize successfully to more complex constraints (SuperClue and CFBench) and out-of-domain constraints (IFBench).

\subsection{Zero-RL Performance}
The Light-IF-1.7B-Zero pipeline starts from the base Qwen3-1.7B model, involving Zero-RL training, thinking-pattern extraction, a cold-start phase, and sequential easy-to-hard RL training. Through this process, we derive four models: 1) Light-IF-1.7B-ZeroRL, directly obtained after Zero-RL. 2) Light-IF-1.7B-EntSFT(ZR), trained from Qwen3-1.7B via extracted samples of Light-IF-1.7B-ZeroRL. 3) Light-IF-1.7B-EntSFT(ZR)-TEARL, obtained by applying TEA-RL to Light-IF-1.7B-EntSFT(ZR) on easy prompts. 4) Light-IF-1.7B-EntSFT(ZR)-TEARL-s2, obtained by TEA-RL to Light-IF-1.7B-EntSFT(ZR)-TEARL on hard prompts.

The results shown in \cref{tab:zero} demonstrate that each step in the pipeline progressively enhances model performance, with the final model achieving the best results overall. Notably, the final Light-IF-1.7B-Zero exhibits instruction-following performance superior to the Qwen3-32B model on IFEval and IFBench, and comparable on SuperClue.

\subsection{Ablation Study}

To validate the effectiveness of each design component within our overall framework, we conduct an ablation study. Specifically, we consider two steps that can potentially be omitted: the cold-start phase and sequential RL phase with dense rewards. Additionally, the two novel components proposed, Entropy-SFT and TEA-RL, can be replaced by existing techniques. To comprehensively assess the contribution of each design, we compare seven models: 1) Qwen3-1.7B: base model. 2) Light-IF-1.7B-EntSFT: model without sequential RL. 3) Light-IF-1.7B-ZeroRL: model without cold-start. 4) Light-IF-1.7B-SFT-GRPO: model with SFT and GRPO. 5) Light-IF-1.7B-EntSFT-GRPO: model with EntropySFT and GRPO. 6) Light-IF-1.7B-SFT-TEARL: model with SFT and TEARL. 7) Light-IF-1.7B-EntSFT-TEARL: model with EntropySFT and TEARL.

The ablation study results presented in \cref{tab:ablation} clearly demonstrate the individual contributions of each step and method. By comparing the performance of models 1, 2, 3, and 7, we conclude that both the cold-start phase and the sequential RL stage significantly enhance the final performance. Furthermore, improvements observed when comparing model 4 with 5 and 6 validate the effectiveness of our proposed Entropy-SFT and TEA-RL over their counterparts, standard SFT and GRPO. Integrating all proposed components, Light-IF-1.7B-EntSFT-TEARL achieves the highest performance across all four benchmarks. Importantly, our results highlight that entropy plays a critical role in our framework—first during cold-start (via Entropy-SFT) and again during RL (via TEA-RL)—and their combined implementation is beneficial for instruction-faithful generation.

\subsection{Entropy Dynamics within the Framework}

Entropy control is one of the most prominent features of our framework. In this section, we demonstrate the effectiveness of our entropy management strategy through two aspects: the dynamics of entropy over training and the evolution of high-entropy tokens at different stages. The entropy dynamics experiment is conducted using RL stage 1 prompts with the 1.7B model, while the token dynamics experiment is conducted on easy prompts in English for readability.

\begin{figure}[t]
  \centering
  \includegraphics[width=0.88\linewidth]{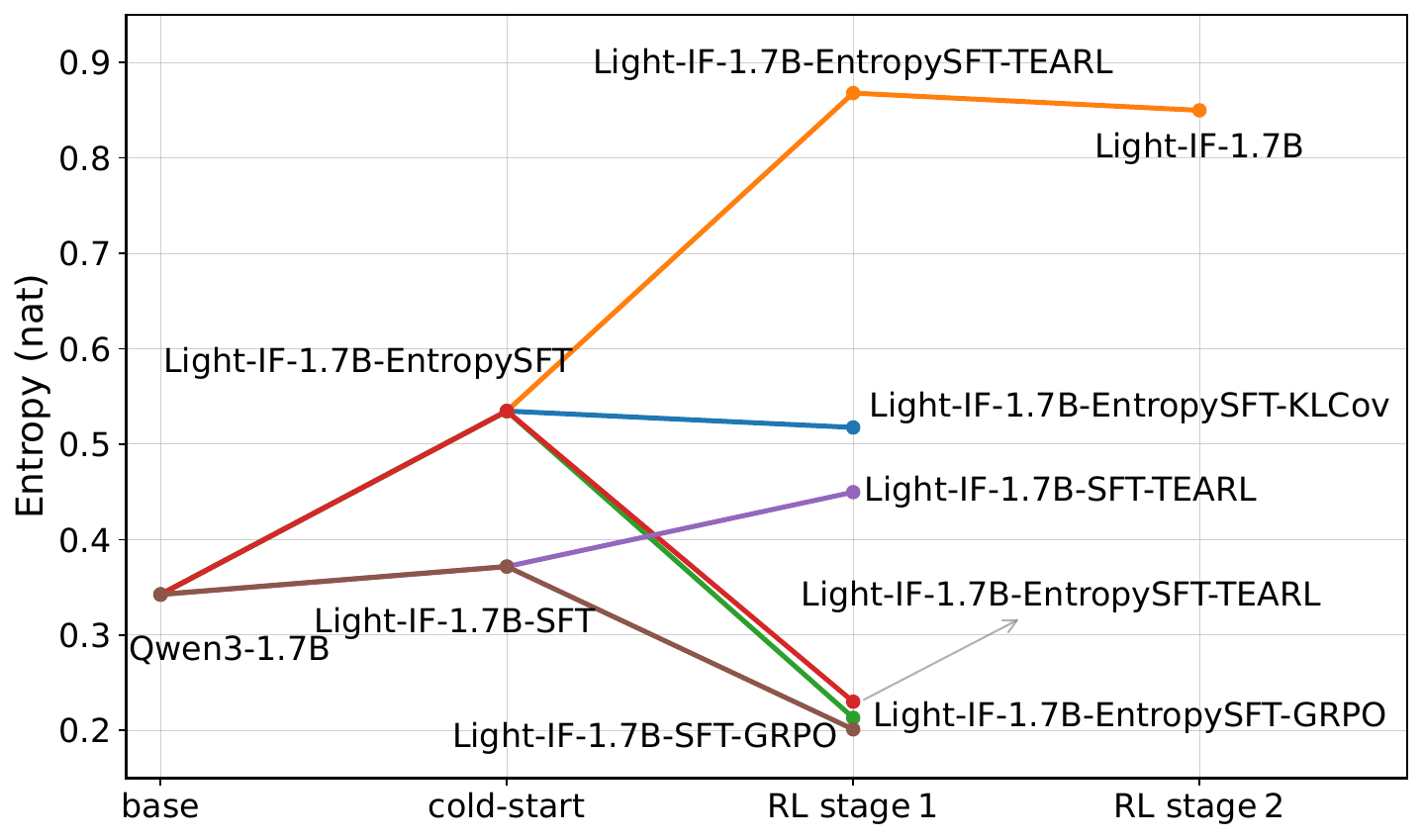}
  \caption{Entropy dynamics. Each point represents a model, and branching paths indicates the adoption of different training strategies in each stage.}
  \label{fig:ent_dyn}
\end{figure}

Firstly, entropy dynamics across training steps are illustrated in \cref{fig:ent_dyn}. Results clearly show that both Entropy-SFT and TEA regularization effectively increase model entropy, promoting exploration. In contrast, standard SFT and conventional RL methods (standard GRPO, GRPO with entropy regularization, KL-Cov~\citep{cui2025entropy}) reduce entropy (detailed analyses and fine-grained entropy dynamics during RL stage 1 are provided in Appendix D). We argue that retaining relatively high entropy at the end of training remains beneficial, as overly deterministic outputs may still produce errors unrelated to high-entropy tokens. Moreover, excessively confident models are less desirable in practical settings. Maintaining higher entropy encourages exploration of alternative reasoning paths, enabling self-checking and reflection, thus enhancing the probability of correct outcomes.

\begin{figure}[t]
  \centering
  \includegraphics[width=1.0\linewidth]{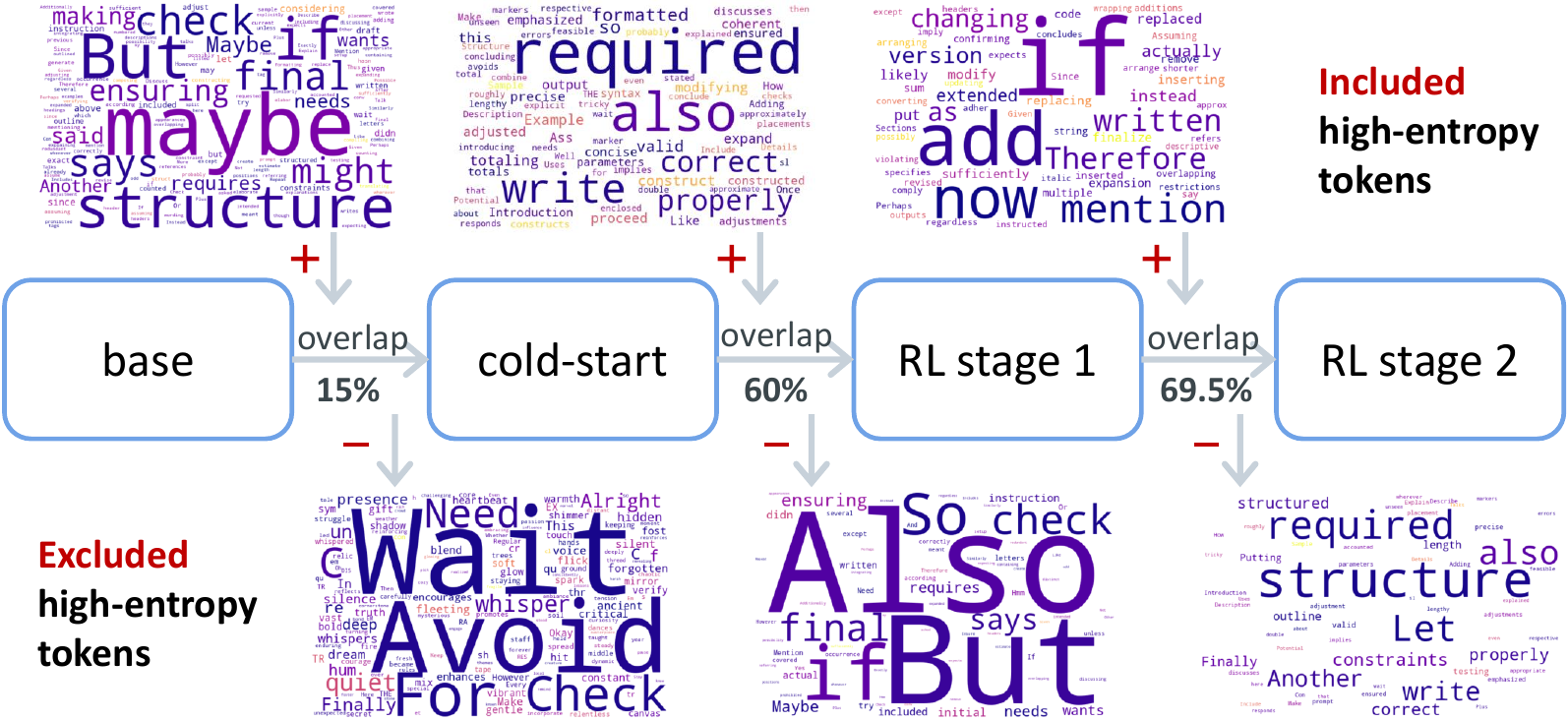}
  \caption{High-entropy token dynamics throughout training. Tokens with Top-200 average entropy and frequency more than 100 are viewed high-entropy.}
  \label{fig:tok_dyn}
\end{figure}

Next, we examine high-entropy token dynamics as shown in \cref{fig:tok_dyn}. Two prominent trends emerge: 1) the overlap of high-entropy tokens between the base and the cold-start models is minimal. Transition words and tokens related to previewing and self-checking replace content-focused verbs and nouns, signaling the emergence of a new reasoning pattern. 2) In later stages, high-entropy token overlap significantly increases, indicating that changes during the RL stage are subtle—only a subset of tokens undergo entropy adjustments while preserving the overall distribution of high-entropy tokens. An intriguing observation is that certain tokens (e.g., \textit{check}) initially become high-entropy tokens and subsequently excluded. The underlying reason is the increased prediction probability. Specifically, for tokens with low initial probability, an increase in prediction probability first raises their entropy and subsequently decreases it. This phenomenon further confirms that tokens like "check" become deeply integrated into the model's reasoning pattern.

\section{Conclusion}

In this paper, we propose a comprehensive framework to equip LLMs with generalizable reasoning for complex instruction following. Specifically, we first design a pipeline to synthesize complex prompts. Then, we post-train the base model via Zero-RL on synthetic prompts to encourage preview and self-checking behaviors. Leveraging responses from the Zero-RL model (optionally with external APIs), we collect effective reasoning samples and conduct Entropy-SFT to cold-start the base model. Subsequently, a two-stage easy-to-hard RL with TEA regularization further enhances the reasoning capabilities. The resulting models, Light-IF-32B, Light-IF-1.7B, and Light-IF-1.7B-Zero, substantially outperform their respective base models. Notably, Light-IF-32B significantly surpasses powerful LLMs such as DeepSeek-R1 and Doubao-1.6, while the comparable performance between Light-IF-1.7B-Zero and Light-IF-1.7B demonstrates the promising solution for eliciting effective reasoning patterns without external guidance.

\bibliography{Light-IF}

\appendix
\clearpage
\section{A. Light Training Cost}
The training cost for the Light-IF series models is relatively lightweight. Specifically, the Light-IF-32B model incurs a training cost of approximately \$2,800, utilizing 11 nodes of A800 $\times$8 GPUs over 30 hours. Training the Light-IF-1.7B model costs around \$342, utilizing 4 nodes of A800$\times$8 GPUs over 10 hours.

\section{B. Implementation Details}
In this section, we provide detailed hyperparameters used during the cold-start stage, RL stage 1, and RL stage 2, as summarized in \cref{tab:parameter}.

\begin{table}[h]
\centering
\footnotesize
\setlength{\tabcolsep}{3pt}
\begin{tabular}{lcccc}
\hline
Model Names        & LR            & Batch Size          & Length     & Steps \\ \hline
\multicolumn{5}{c}{SFT}                                                       \\ \hline
Light-IF-1.7B-Zero & 1e-5          & 16                  & 16K        & 375   \\
Light-IF-1.7B      & 1e-5          & 16                  & 16K        & 375   \\
Light-IF-32B       & 5e-6          & 32                  & 16K        & 187   \\ \hline
\multicolumn{5}{c}{RL Stage1}                                                 \\ \hline
Light-IF-1.7B-Zero & 5e-6          & 128 ($n$=8, mini=32)  & 8K         & 100   \\
Light-IF-1.7B      & 5e-6          & 128 ($n$=8, mini=32)  & 8K         & 100   \\
Light-IF-32B       & 5e-6          & 352 ($n$=8, mini=88)  & 8K         & 100   \\ \hline
\multicolumn{5}{c}{RL Stage2}                                                 \\ \hline
Light-IF-1.7B-Zero & 5e-6          & 128 ($n$=16, mini=32) & 16K        & 30    \\
Light-IF-1.7B      & 5e-6          & 128 ($n$=16, mini=32) & 16K        & 30    \\
Light-IF-32B       & 5e-6          & 352 ($n$=16, mini=88) & 16K        & 75    \\ \hline
\end{tabular}
\caption{Training Hyperparameters. $n$ refers to $n$ responses per prompt during rollout, mini refers to mini batch size during RL policy optimization.}
\label{tab:parameter}
\end{table}

\section{C. Cold-Start Data Filter}
The filtering pipeline comprises three steps: correctness check, thinking check, and fluency check. In the correctness check step, we collect samples that strictly follow instructions, verified by automated code-based checks. In the subsequent thinking check step, we filter out samples exhibiting potentially lazy thinking patterns. Specifically, we apply a heuristic rule assuming that shorter thinking content generally indicates less thorough reasoning, discarding samples with fewer than 1,000 tokens. Finally, we remove samples with disfluent expressions by employing an LLM (ChatGPT-4o) as the evaluator. The prompt used for the LLM is provided in \cref{tab:template}, and samples scoring below 8 are filtered out. After these steps, we select the top 2,000 samples ranked by reasoning length from the remaining candidates to form the final cold-start dataset.

\section{D. Entropy Control during RL}
We propose the Token-wise Entropy-Adaptive (TEA) method to maintain sufficient entropy during the RL stage. In this section, we compare TEA with existing entropy control methods from the literature, including entropy regularization and KL-Cov~\citep{cui2025entropy}.

\begin{figure}[ht]
  \centering
  \includegraphics[width=0.95\linewidth]{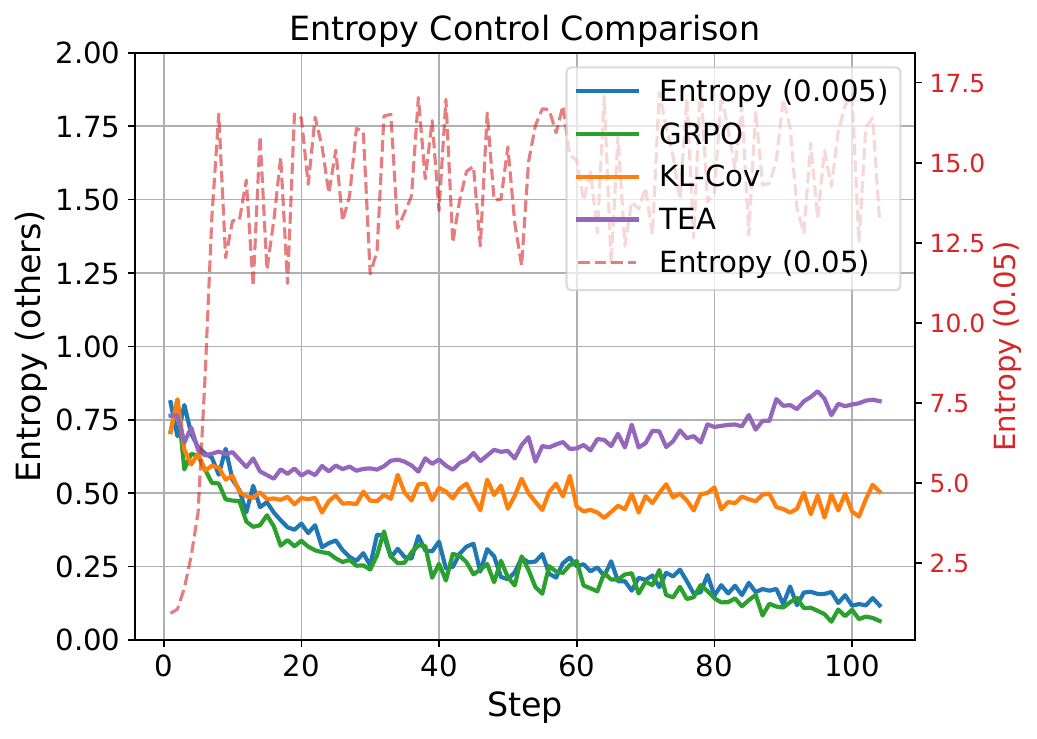}
  \caption{Entropy control during RL.}
  \label{fig:entropy}
\end{figure}

Experiments are conducted using the Light-IF-1.7B-EntSFT model with different entropy control strategies. The results presented in \cref{tab:rl} indicate that TEA-RL consistently achieves the best performance across all four benchmarks. Additionally, the entropy dynamics over training steps are illustrated in \cref{fig:entropy}. It is clear from the figure that TEA effectively maintains a high entropy level, whereas traditional entropy regularization methods tend to either cause entropy collapse or entropy explosion. Notably, despite sharing a similar theoretical foundation with KL-Cov, our token-wise entropy-adaptive regularization method demonstrates superior performance.

\section{E. Dense Reward versus Sparse Reward}
The multi-constraint nature of complex instruction-following tasks motivates the design of the dense reward. In this section, we compare the appearances of dense and sparse reward. Given the differences in absolute values between dense and sparse rewards, we normalize reward scores by dividing by the maximum possible reward value to obtain a relative reward score. The experiments are conducted using Light-IF-1.7B-EntSFT with TEA-RL.

\begin{figure}[h]
  \centering
  \includegraphics[width=1.0\linewidth]{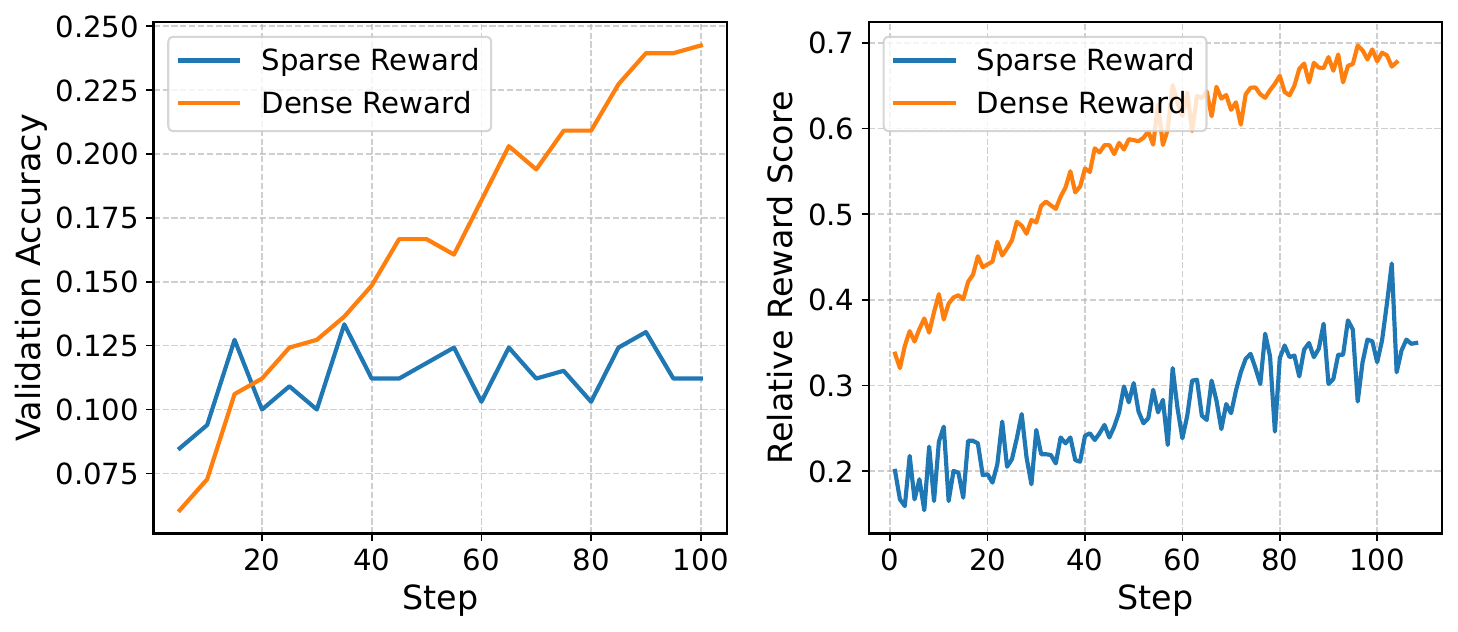}
  \caption{Appearances of the dense and sparse reward.}
  \label{fig:reward}
\end{figure}

As illustrated in \cref{fig:reward}, RL with sparse rewards exhibits slow improvement in reward scores, stagnates in validation accuracy, and displays considerable fluctuations. Conversely, RL with dense rewards demonstrates smooth and rapid improvements in both validation accuracy and relative reward scores, clearly indicating superior performance.

\begin{table*}[t]
\centering
\small
\setlength{\tabcolsep}{3pt}       
\begin{tabular}{lccccccccccccc}
\hline
\multirow{2}{*}{Method} & \multirow{2}{*}{Superclue} & \multicolumn{5}{c}{IFEval}                                                & \multicolumn{4}{c}{CFBench}                          & \multicolumn{3}{c}{IFBench}             \\
                                &                                     & LP             & LI             & SP             & SI             & AVG            & CSR           & ISR           & PSR           & AVG           & PL             & IL             & AVG            \\ \hline
GRPO                            & 0.153                               & 0.815          & 0.872          & 0.776          & 0.842          & 0.826          & \textbf{0.85} & 0.59          & \textbf{0.70} & 0.71          & 0.323          & 0.361          & 0.342          \\
Entropy (0.005)                 & 0.162                               & 0.843          & 0.892          & 0.786          & 0.853          & 0.844          & \textbf{0.85} & \textbf{0.60} & \textbf{0.70} & \textbf{0.72} & 0.313          & 0.349          & 0.331          \\
Entropy (0.05)                  & -                                   & -              & -              & -              & -              & -              & -             & -             & -             & -             & -              & -              & -              \\
KL-Cov                         & 0.209                               & 0.861          & 0.903          & 0.826          & 0.878          & 0.867          & 0.83          & 0.57          & 0.68          & 0.69          & \textbf{0.347} & 0.388          & \textbf{0.368} \\
TEA                             & \textbf{0.231}                      & \textbf{0.880} & \textbf{0.917} & \textbf{0.845} & \textbf{0.891} & \textbf{0.883} & \textbf{0.85} & \textbf{0.60} & \textbf{0.70} & \textbf{0.72} & 0.344          & \textbf{0.391} & \textbf{0.368} \\ \hline
\end{tabular}
  \caption{Comparisons of Entropy control strategies. LP: loose prompt, LI: loss instruction, SP: strict prompt, SI: strict instruction, PL: prompt level, IL: instruction level. Best are marked \textbf{bold}.}
  \label{tab:rl}
\end{table*}

\begin{table*}[t]
    \centering
    \begin{tabular}{p{0.95\linewidth}}
    \toprule
    Please act as a reviewer and evaluate the quality of the model's responses as an AI assistant. \\
    Your evaluation should prioritize the \textbf{fluency and readability} of the final answer. If the response is crafted solely to meet specific constraints and sacrifices natural flow or clarity, it should be rated poorly. For example, if the instruction requires the letter ``n'' to appear at least three times, and the response artificially includes multiple standalone instances of the letter ``n'' just to fulfill this requirement, such a response should receive a low score. \\
    Next, consider whether the response \textbf{fulfills the instruction requirements}, and evaluate it strictly according to the \textbf{scoring criteria below}: \\
    1) Score: 1--2, \textbf{Criteria:} The response is of poor quality, lacks fluency, and is written solely to satisfy the instruction constraints without regard for the overall quality of the reply. \\
    2) Score: 3--5, \textbf{Criteria:} The response is generally readable but lacks overall fluency and smoothness. \\
    3) Score: 6--8, \textbf{Criteria:} The response is of relatively high quality, with generally fluent and coherent language. \\
    4) Score: 9--10, \textbf{Criteria:} The response is of high quality, with fluent, natural, and well-structured language. \\
    Please provide the following: 1) A brief explanation to evaluate the quality of the AI assistant's response. If there are any issues with response quality or language fluency, please identify and briefly explain them; \\
    2) Then provide an evaluation score, which must be strictly graded according to the following format: ``[[rating]]'', example: ``Score:[[4]]''. \\
    \texttt{[Question]} \\
    \textit{\textless question\textgreater{}} \\
    \texttt{[The Start of Assistant's Answer]} \\
    \textit{\textless answer\textgreater{}} \\
    \texttt{[The End of Assistant's Answer]} \\
    \bottomrule
    \end{tabular}
    \caption{Template for fluency check. Placeholders \textit{\textless question\textgreater{}} and \textit{\textless answer\textgreater{}} will be replaced with actual inputs.}
    \label{tab:template}
\end{table*}

\end{document}